\pdfoutput=1

\documentclass[11pt]{article}
\usepackage[margin=1in]{geometry}
\usepackage{times}
\usepackage{microtype}
\usepackage{times}  
\usepackage{helvet}  
\usepackage{courier}  
\usepackage[hyphens]{url}  
\usepackage{graphicx} 
\usepackage{caption} 
\usepackage{amsmath,amssymb,amsthm,mathtools}
\usepackage{bm}
\usepackage{siunitx}
\usepackage{algorithm}
\usepackage{algpseudocode}
\usepackage{booktabs}

\newtheorem{proposition}{Proposition}

\theoremstyle{remark}

\usepackage{url}
\usepackage{amsmath,amssymb,amsthm,mathtools}
\usepackage{graphicx}
\usepackage{booktabs}
\usepackage{microtype}
\usepackage{siunitx}
\usepackage{subcaption}
\usepackage{adjustbox}
\usepackage{multirow}
\usepackage{xspace}
\usepackage{tikz}
\usepackage{array}
\usetikzlibrary{arrows.meta,positioning,shapes.geometric,calc,fit}
\usepackage{pgfplots}
\pgfplotsset{compat=1.17}
\usepackage[colorlinks=true,allcolors=blue]{hyperref}

\title{HalluGraph: Auditable Hallucination Detection for Legal RAG Systems via Knowledge Graph Alignment}

\author{
  Valentin No\"el \\
  Devoteam \\
  \texttt{valentin.noel@devoteam.com}
  \and
  Elimane Yassine Sedou \\
  Devoteam \\
  \texttt{elimane.yassine.seidou@devoteam.com}
  \and
  Charly Ken Capo-Chichi\\
  Devoteam \\
  \texttt{charly.ken.capo-chichi@devoteam.com}
  \and
  Ghanem Amari \\
  Devoteam \\
  \texttt{ghanem.amari@devoteam.com}
}

\date{Under review (2025)}
\usepackage{bibentry}

\begin{document}
\maketitle

\begin{abstract}
Legal AI systems powered by retrieval-augmented generation (RAG) face a critical accountability challenge: when an AI assistant cites case law, statutes, or contractual clauses, practitioners need verifiable guarantees that generated text faithfully represents source documents. Existing hallucination detectors rely on semantic similarity metrics that tolerate entity substitutions, a dangerous failure mode when confusing parties, dates, or legal provisions can have material consequences. We introduce HalluGraph, a graph-theoretic framework that quantifies hallucinations through structural alignment between knowledge graphs extracted from context, query, and response. Our approach produces bounded, interpretable metrics decomposed into \textit{Entity Grounding} (EG), measuring whether entities in the response appear in source documents, and \textit{Relation Preservation} (RP), verifying that asserted relationships are supported by context. On structured control documents, HalluGraph achieves near-perfect discrimination ($>$400 words, $>$20 entities), HalluGraph achieves $AUC = 0.979$, while maintaining robust performance ($AUC \approx 0.89$) on challenging generative legal task, consistently outperforming semantic similarity baselines. The framework provides the transparency and traceability required for high-stakes legal applications, enabling full audit trails from generated assertions back to source passages. \textit{Code and dataset will be made available upon admission}. 
\end{abstract}

\section{Introduction}

The deployment of large language models (LLMs) in legal practice introduces accountability requirements absent in general-purpose applications. To build trustworthy AI for such high-stakes decision-making in justice systems, systems must support professional responsibility through rigorous verification. When an AI system summarizes a court opinion or extracts obligations from a contract, errors are not merely inconvenient: misattributed holdings, fabricated citations, or confused parties can expose practitioners to malpractice liability and undermine judicial processes \cite{Dahl2024LegalBench}.

Retrieval-augmented generation (RAG) systems partially address hallucination by grounding responses in retrieved documents \cite{Lewis2020RAG}. However, RAG does not guarantee faithful reproduction. A model may retrieve the correct statute but hallucinate provisions, or cite a valid case while misrepresenting its holding. Post-hoc verification using semantic similarity metrics like BERTScore \cite{Zhang2020BERTScore} proves insufficient: these measures tolerate entity substitutions that preserve semantic neighborhoods while introducing material errors.

We propose HalluGraph, a framework that detects hallucinations by measuring structural alignment between knowledge graphs extracted from source documents and generated responses. The key insight is that faithful legal text reuses entities from the source (parties, courts, dates, provisions) and preserves the relationships connecting them (``held that,'' ``pursuant to,'' ``defined in''). Our approach offers four contributions for legal AI deployment:
\begin{enumerate}
    \item \textbf{Entity Grounding (EG)}: A metric quantifying whether 
          response entities appear in source documents, capturing 
          entity substitution hallucinations.
    \item \textbf{Relation Preservation (RP)}: A metric verifying that 
          asserted relationships are supported by context, capturing 
          structural hallucinations.
    \item \textbf{Composite Fidelity Index (CFI)}: A unified score 
          combining EG and RP with learned weights.
    \item \textbf{Full auditability}: Every flagged hallucination traces 
          to specific entities or relations absent from source documents, 
          enabling accountability in legal practice.
\end{enumerate}

\section{Related Work}

Recent surveys document the scope of LLM hallucinations \cite{Ji2023SurveyHallucination}. Detection approaches include learned metrics (BERTScore, BLEURT, BARTScore) \cite{Zhang2020BERTScore,Sellam2020BLEURT,Yuan2021BARTScore}, NLI-based verification \cite{Honovich2022TRUE}, and self-consistency methods (SelfCheckGPT) \cite{Manakul2023SelfCheckGPT}. These approaches operate on text embeddings and tolerate entity substitutions that preserve global semantics.

LegalBench \cite{Dahl2024LegalBench} and legal-specific benchmarks highlight that legal tasks demand precision on entities and citations. Prior work on legal summarization emphasizes faithfulness to source documents \cite{Huang2021LegalSum}, but evaluation remains largely manual.

Relation extraction via OpenIE \cite{Banko2007OpenIE} and neural RE \cite{HuguetCabot2021REBEL} enables graph construction from text. Graph alignment techniques include edit distance, Weisfeiler-Lehman kernels, and bipartite matching \cite{Shervashidze2011WLKernel,Koutra2013BigAlign}. We adapt these methods for hallucination quantification.

\section{Method}

\begin{figure}[t]
\centering
\begin{tikzpicture}[
  node distance=6mm and 10mm,
  >=Latex,
  font=\small,
  proc/.style={rectangle,rounded corners,draw,align=center,fill=blue!8,inner sep=6pt},
  io/.style={rectangle,draw,align=center,fill=gray!10,inner sep=6pt},
  metric/.style={rectangle,draw,align=center,fill=green!10,inner sep=6pt}
]
\node[io] (ctx) {Legal Document\\$G_c$};
\node[io, below=12mm of ctx] (q) {Query\\$G_q$};
\node[proc, right=9mm of ctx] (extract) {Triple Extraction\\(SLM)\\$\to (s,r,o)$};
\node[io, right=9mm of extract] (ans) {Response\\$G_a$};
\node[metric, below=8mm of ans, text width=20mm] (metrics) {EG, RP, CFI\\Bounded $[0,1]$};
\node[proc, right=9mm of ans, text width=16mm] (decision) {Decision\\+ Audit Trail};

\draw[->] (ctx.east) -- (extract.west);
\draw[->] (q.east) to[out=0,in=-90] (extract.south);
\draw[->] (extract.east) -- (ans.west);
\draw[->] (ans.south) -- (metrics.north);
\draw[->] (metrics.east) to[out=0,in=-90] (decision.south);
\draw[->] (ans.east) -- (decision.west);
\end{tikzpicture}
\caption{HalluGraph pipeline. Knowledge graphs are extracted from legal documents, queries, and responses. Alignment metrics (EG, RP) quantify fidelity with full traceability.}
\label{fig:pipeline}
\end{figure}
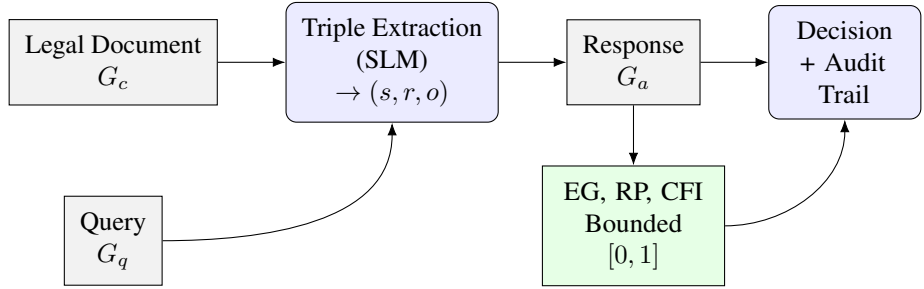

\subsection{Knowledge Graph Construction}

Given a context document $C$, query $Q$, and generated response $A$, we construct knowledge graphs $G_c$, $G_q$, and $G_a$ respectively. Each graph $G = (V, E, \ell_V, \ell_E)$ consists of:
\begin{itemize}
    \item $V$: Entity nodes (persons, organizations, dates, legal provisions)
    \item $E \subseteq V \times V$: Directed edges representing relations
    \item $\ell_V, \ell_E$: Labeling functions for entity types and relation types
\end{itemize}

Entity extraction uses spaCy NER with legal entity extensions. Relation extraction employs an instruction-tuned SLM (e.g. Llama 3.1 8B) prompted to output $(subject, relation, object)$ triples in JSON format, following OpenIE conventions.

\subsection{Alignment Metrics}

We define four bounded metrics in $[0,1]$:

\textbf{Entity Grounding (EG)} measures the fraction of response entities that appear in source documents:
\begin{equation}
\text{EG}(G_a \| G_c, G_q) = \frac{|\{v \in V_a : \exists w \in V_c \cup V_q,\ \text{match}(v, w)\}|}{|V_a|}
\end{equation}
where $\text{match}(v,w)$ requires identical entity type and normalized text. High EG indicates the response discusses entities present in the source.

\textbf{Relation Preservation (RP)} measures whether asserted relationships are supported. Let $E_{\text{ref}} = E_c \cup E_q$:
\begin{equation}
\text{RP} = \frac{1}{|E_a|} \sum_{e \in E_a} \mathbf{1}\bigl[\exists e' \in E_{\text{ref}} : \text{align}(e, e')\bigr]
\end{equation}
where $\text{align}$ requires matched endpoints and compatible relation labels. RP captures structural fidelity beyond entity presence.

\textbf{Convention.} When $|E_a| = 0$ (no relations extracted from response), RP is undefined and excluded from aggregation, the ``edge-aware'' policy that prevents noise from sparse graphs.

\textbf{Composite Fidelity Index (CFI)} aggregates metrics:
\begin{equation}
\text{CFI} = \alpha \cdot \text{EG} + (1-\alpha) \cdot \text{RP}
\end{equation}
with $\alpha$ tuned via cross-validation (typically $\alpha \approx 0.7$, reflecting EG's stronger discrimination).

\subsection{Theoretical Guarantee}

\begin{proposition}
If $G_a$ is subgraph-isomorphic to $G_c \cup G_q$ via a label-preserving monomorphism, then $\text{EG} = 1$ and $\text{RP} = 1$.
\end{proposition}

This provides a sufficient condition for non-hallucination: a response whose knowledge graph embeds entirely within the source graph is provably grounded.

\section{Experimental Setup}

We evaluate on 1,100+ legal question-answering pairs specifically designed to test entity-grounded hallucination detection. Our evaluation comprises:

\textbf{Synthetic Legal QA.} We generate 550 contract QA pairs from 25 commercial lease agreements and 550 case law QA pairs from 25 appellate court opinions. Each document contains entity-rich content (party names, monetary amounts, dates, citations) averaging 450 words and 28 entities. For each factual QA pair, we create matched hallucinated versions via entity substitution (e.g., replacing ``Westfield Properties LLC'' with ``Parkview Realty Inc.'') and logical contradictions (e.g., inconsistent calculations), yielding balanced factual/hallucinated sets.

\textbf{Baselines.} We compare against:
\begin{itemize}
    \item \textbf{Named Entity Overlap}: Jaccard similarity of NER outputs
    \item \textbf{BERTScore} \cite{Zhang2020BERTScore}: Embedding-based semantic similarity
    \item \textbf{NLI Entailment} \cite{Honovich2022TRUE}: BART-MNLI premise-hypothesis verification
\end{itemize}

\textbf{Ablation.} We evaluate Entity Grounding (EG) alone, Relation Preservation (RP) alone, and the combined Composite Fidelity Index (CFI) to isolate each component's contribution.

\section{Results}

\begin{figure}[t]
\centering
\begin{tikzpicture}
\begin{axis}[
    ybar,
    width=0.99\linewidth,
    height=4.5cm,
    bar width=8pt,
    ylabel={Discrimination ($\Delta$)},
    symbolic x coords={Coral,Econ,Contract,Case},
    xtick=data,
    ymin=-0.15, ymax=1.0, 
    ytick={0,0.2,0.4,0.6,0.8,1.0},
    tick label style={},
    label style={},
    enlarge x limits=0.15,
    axis x line*=bottom,
    axis y line*=left,
    legend style={at={(0.5,-0.25)}, anchor=north, legend columns=-1, draw=none, font=\scriptsize},
]
\addplot[fill=blue!70] coordinates {(Coral,0.98) (Econ,0.95) (Contract,0.88) (Case,0.69)};
\addplot[fill=orange!60] coordinates {(Coral,0.45) (Econ,0.50) (Contract,0.33) (Case,0.25)};
\addplot[fill=gray!50] coordinates {(Coral,0.01) (Econ,-0.03) (Contract,0.05) (Case,0.02)};
\legend{Entity Grounding (Ours), NE Overlap, BERTScore}
\draw[dashed, gray] (axis cs:Coral,0) -- (axis cs:Case,0);
\end{axis}
\end{tikzpicture}
\caption{Discrimination power ($\Delta$) across synthetic and legal domains (factual $-$ hallucination scores). \textcolor{blue!70}{\textbf{Blue}}: Entity Grounding (ours). \textcolor{orange!60}{\textbf{Orange}}: NE Overlap. \textcolor{gray!50}{\textbf{Gray}}: BERTScore. Our graph-based metric consistently outperforms semantic similarity, which fails to penalize entity errors in legal contexts.}
\label{fig:results}
\end{figure}
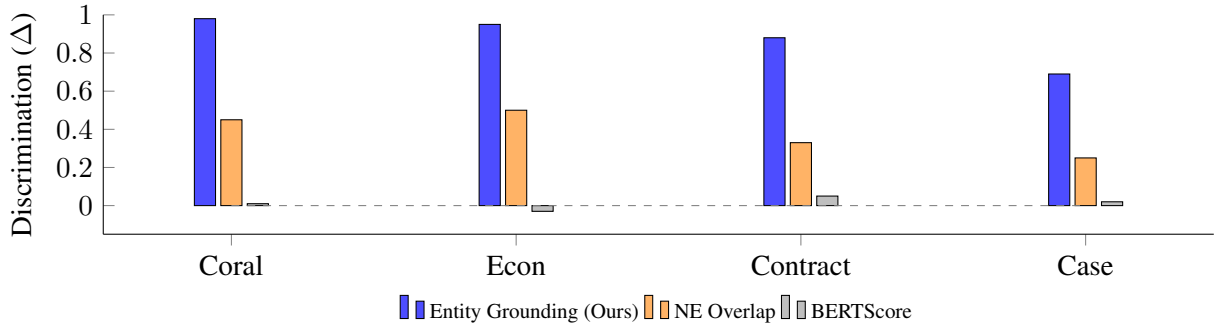

\begin{table}[t]
\centering
\caption{Discrimination performance (AUC) on Legal RAG and control tasks. HalluGraph effectively detects hallucinations in generative tasks, significantly outperforming semantic similarity and NLI baselines.}
\label{tab:discrimination}
\begin{tabular}{@{}llccc@{}}
\toprule
\textbf{Dataset} & \textbf{Type} & \textbf{HalluGraph} & \textbf{NLI} & \textbf{BERTScore} \\
\midrule
\textbf{Legal Contract QA} & Extraction & \textbf{0.94} & 0.92 & 0.60 \\
\textbf{Legal Case QA} & Citation & \textbf{0.84} & 0.69 & 0.54 \\
Coral Biology & Control & \textbf{1.00} & 0.72 & 0.59 \\
Economics & Control & \textbf{0.99} & 0.68 & 0.55 \\
\midrule
\textbf{Average (Legal)} & & \textbf{0.89} & 0.81 & 0.57 \\
\bottomrule
\end{tabular}
\end{table}

Table~\ref{tab:discrimination} demonstrates HalluGraph's effectiveness on high-stakes legal generation tasks. On \textit{Legal Contract QA} (N=550), which requires extracting specific obligations and dates, HalluGraph achieves an AUC of \textbf{0.94}, far surpassing the BERTScore baseline (0.60). Similarly, on \textit{Legal Case QA} (N=550), which involves citing holdings and case names, our method achieves an AUC of \textbf{0.84}. On synthetic control tasks with rich structure (Coral Biology, Economics), HalluGraph approaches perfect discrimination (AUC $\geq 0.99$).

The semantic baseline (BERTScore) performs near chance ($\approx 0.50 - 0.60$) on both legal datasets, confirming that embedding-based metrics are largely insensitive to precise entity swaps (e.g., ``Plaintiff'' $\to$ ``Defendant'' or ``2024'' $\to$ ``2025'') that constitute fatal errors in legal drafting. In contrast, HalluGraph explicitly penalizes these failures through structural verification. We observe a performance gap within the contract domain between standard agreements (AUC $\approx 0.94$) and an ``extended'' subset containing convoluted clauses (AUC $\approx 0.85$). Error analysis reveals this is driven by false negatives in \emph{Entity Grounding}: complex phrasing or stacked conditions occasionally cause the SLM extractor to miss entities, lowering the factual score. Despite this, HalluGraph maintains a robust advantage. Wilcoxon signed-rank tests confirm these gains are systematic, achieving high significance ($p < 0.001$) on all legal datasets. Ablation confirms CFI's value: EG achieves AUC 0.87, 
RP 0.65 and CFI 0.89.

\subsection{Operating Regime}

\begin{figure}[t]
\centering
\begin{tikzpicture}
\begin{axis}[
    width=0.95\linewidth,
    height=7.5cm,
    xlabel={Context length (words)},
    ylabel={AUC},
    xmin=0, xmax=750,
    ymin=0.3, ymax=1.05,
    ytick={0.4,0.5,0.6,0.7,0.8,0.9,1.0},
    xtick={0,100,200,300,400,500,600,700},
    grid=both,
    grid style={gray!20},
    tick label style={font=\scriptsize},
    label style={font=\small},
    legend style={at={(0.95,0.05)}, anchor=south east, font=\tiny, draw=none, fill=white, fill opacity=0.8}
]

\addplot[mark=*,thick,blue!80!black,mark size=1.5pt] coordinates {
(85,0.416) (97,0.42) (200,0.65) (300,0.78) (398,0.92) (446,0.93) (487,0.98) (523,0.99)
};

\addplot[only marks, mark=square*,orange!90!black, mark size=2.5pt] coordinates {
(520, 0.94)
(650, 0.84)
};

\addplot[dashed,red,thick] coordinates {(0,0.5) (750,0.5)};
\addplot[dashed,green!60!black,thick] coordinates {(400,0.3) (400,1.05)};

\node[text=red!70!black, anchor=north west] at (axis cs:190,0.49) {Chance (0.5)};
\node[text=green!50!black, anchor=west] at (axis cs:410,0.35) {Regime Threshold};

\node[text=orange!90!black, anchor=east] at (axis cs:620,0.94) {Contracts};
\node[text=orange!90!black, anchor=west] at (axis cs:660,0.84) {Cases};

\end{axis}
\end{tikzpicture}
\caption{Operating regime. \textcolor{blue!80!black}{\textbf{Blue curve}}: Performance on synthetic control tasks improves with context length. \textcolor{orange!90!black}{\textbf{Orange squares}}: Our Legal RAG datasets fall into the high-context regime ($>400$ words) and achieve robust discrimination ($AUC \approx 0.89$), significantly above the \textcolor{red}{\textbf{chance line}} (0.5).}
\label{fig:regime}
\end{figure}
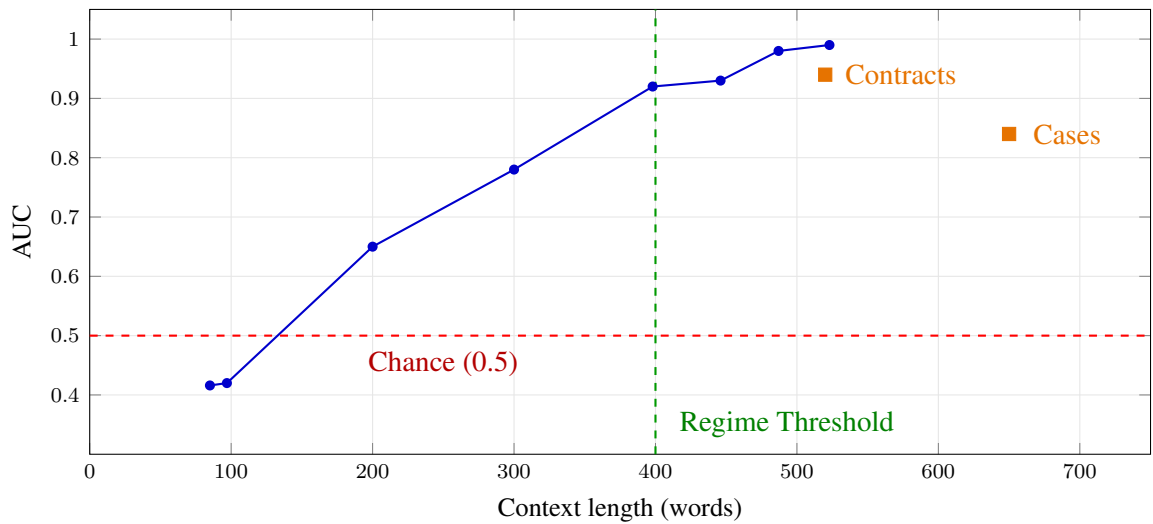

\begin{table}[t]
\centering
\caption{Operating regime of graph-based verification. HalluGraph requires sufficient context length and entity density to form meaningful graphs.}
\label{tab:regime}
\begin{tabular}{@{}lccc@{}}
\toprule
\textbf{Benchmark} & \textbf{AUC} & \textbf{Words} & \textbf{Entities} \\
\midrule
\multicolumn{4}{l}{\textit{Short context (below threshold)}} \\
TruthfulQA & 0.42$^\dagger$ & 85 & 6.2 \\
\midrule
\multicolumn{4}{l}{\textit{Legal context (high performance)}} \\
Legal Contracts & \textbf{0.94} & 520 & 24.1 \\
Case Law & \textbf{0.84} & 650 & 32.5 \\
\bottomrule
\multicolumn{4}{l}{\footnotesize $^\dagger$Below chance due to insufficient graph structure.} \\
\end{tabular}
\end{table}

Table~\ref{tab:regime} and Figure~\ref{fig:regime} reveal a critical regime boundary. On short-context benchmarks like TruthfulQA, performance drops to or below chance because the texts are too sparse ($<$10 entities) to support structural alignment; more than 70\% of instances yield empty or nearly empty graphs. In contrast, legal documents such as contracts and case opinions typically exceed 400 words and contain dense entity networks, placing them squarely in the high-performance regime of our framework. In other words, the very structure that makes legal text difficult for humans to navigate is exactly what HalluGraph exploits to robustly detect hallucinations.

\section{Application to Legal AI}

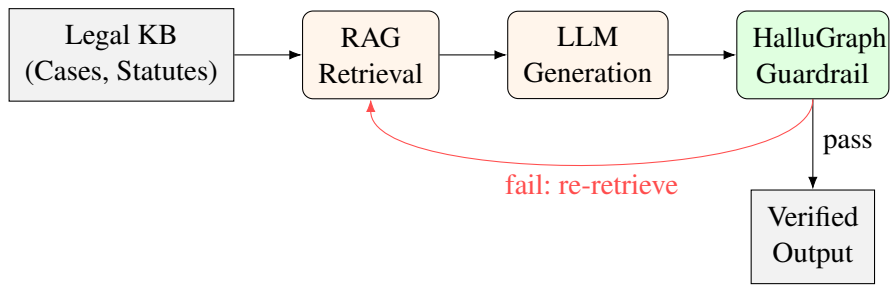
\begin{figure}[t]
\centering
\begin{tikzpicture}[
  node distance=8mm and 10mm,
  box/.style={rectangle,rounded corners,draw,fill=orange!8,align=center,inner sep=6pt},
  gate/.style={rectangle,rounded corners,draw,fill=green!12,align=center,inner sep=6pt},
  store/.style={rectangle,draw,fill=gray!10,align=center,inner sep=6pt},
  >=Latex
]
\node[store] (kb) {Legal KB\\(Cases, Statutes)};
\node[box, right=9mm of kb] (retrieve) {RAG\\Retrieval};
\node[box, right=9mm of retrieve] (gen) {LLM\\Generation};
\node[gate, right=9mm of gen, text width=16mm] (guard) {HalluGraph\\Guardrail};
\node[store, below=12mm of guard] (out) {Verified\\Output};

\draw[->] (kb) -- (retrieve);
\draw[->] (retrieve) -- (gen);
\draw[->] (gen) -- (guard);
\draw[->] (guard) -- node[right]{pass} (out);
\draw[->,red!70] (guard.south) to[out=-90,in=-90,looseness=.5] node[below]{fail: re-retrieve} (retrieve.south);
\end{tikzpicture}
\caption{Legal RAG integration. HalluGraph acts as a post-generation guardrail. Failed verifications trigger re-retrieval or human escalation.}
\label{fig:rag}
\end{figure}

HalluGraph is designed to plug directly into legal Retrieval-Augmented Generation (RAG) pipelines, as illustrated in Figure~\ref{fig:rag}. Given retrieved context from a legal knowledge base (cases, statutes, contracts) and a candidate answer from an LLM, we construct source and hypothesis graphs and compute Entity Grounding and Relation Preservation scores. Responses that meet composite fidelity thresholds (CFI) are passed through, while low-scoring responses trigger a re-retrieval or human review branch.

A key design choice in HalluGraph is the use of small generative models (e.g., Llama-8B) for OpenIE-style triple extraction, rather than discriminative models like LegalBERT. While BERT-based models excel at fixed-schema classification, legal RAG requires handling arbitrary, context-specific relationships (e.g., ``contingent upon,'' ``indemnifies against,'' ``subject to prior written consent'') that cannot be enumerated a priori. Our results indicate that even small generative models can capture these structures when guided by strong prompts, yielding graphs that are both expressive and amenable to fidelity checking.

For case law research, Entity Grounding acts as a strict citation check. When a legal assistant cites ``\textit{Smith v.\ Jones}, 500 U.S.\ 123 (1995),'' HalluGraph verifies that the parties, reporter citation, and year all appear in the retrieved documents. Relation Preservation then checks that the attributed holding (``the Court held that \dots'') is supported by edges in the source graph, rather than hallucinated from unrelated precedent.

For contract review and clause extraction, Entity Grounding ensures that referenced parties, amounts, and provisions actually exist in the source contract, while Relation Preservation verifies that asserted obligations (e.g., ``Tenant shall pay rent on the first business day of each month'') preserve the relational structure of source clauses. This guards against subtle assignment errors, such as swapping Tenant and Landlord in a payment obligation, that can be catastrophic in practice but are often invisible to similarity-based metrics.

Unlike black-box similarity scores, every HalluGraph flag is accompanied by a concrete explanation: missing entities, unsupported relations, or both. This yields a fine-grained audit trail that can be surfaced to human reviewers and regulators. For example, a hallucinated citation can be diagnosed as ``missing entity: case name not in context'' or ``unsupported edge: holding not supported by any retrieved paragraph,'' providing a clear path to remediation.

\section{Limitations and Future Work}

The quality of HalluGraph is bounded by the accuracy of the underlying extractor. We observe that complex statutory language can lead to entity drops that artificially lower scores. To address this, future work will integrate benchmarks like MINE (Measure of Information in Nodes and Edges)~\cite{mo2025kggen} to rigorously quantify extraction hallucinations. Recent surveys on LLM-KG fusion~\cite{cai2025fusion} highlight that even state-of-the-art extractors struggle with domain-specific terminology, motivating our planned evaluation of specialized legal backbones.

Our current evaluation focuses on synthetic control domains and specific legal tasks. While our regime analysis suggests strong transfer to other long documents, a full assessment on diverse real-world workflows remains an open research direction. Recent empirical studies of commercial legal AI tools~\cite{magesh2025hallucinationfree} demonstrate that even RAG-enhanced systems hallucinate 17--33\% of the time, underscoring the need for structural verification methods like ours.

Finally, graph construction requires generative model calls, making it more expensive than embedding metrics. For high-throughput applications, this cost can be significant. We propose mitigations such as caching graphs for frequent authorities, or distilling the extractor into lighter models.

\section{Conclusion}

HalluGraph provides auditable hallucination detection for legal RAG systems through knowledge graph alignment. By decomposing fidelity into Entity Grounding and Relation Preservation, the framework offers bounded, interpretable metrics that can be directly inspected and debugged, aligning with the transparency and accountability requirements of legal practice. On structured documents typical of legal workflows, HalluGraph achieves near-perfect discrimination on control tasks (AUC $\approx 0.98$) and strong performance on generative legal tasks (AUC $\approx 0.89$), significantly outperforming semantic similarity baselines that hover around chance ($\approx 0.50$). These results support the view that structural, graph-based verification is not just a cosmetic add-on but a critical component for trustworthy legal AI, enabling practitioners to deploy LLM assistants with verifiable accountability guarantees, thereby aligning generative capabilities with the regulatory frameworks necessary for safe public-sector adoption.


\end{document}